\documentclass[conference]{IEEEtran}
\IEEEoverridecommandlockouts
\newtheorem{definition}{Definition}
\usepackage{color}

\usepackage{diagbox}

\usepackage{longtable}
\usepackage[T1]{fontenc}% optional T1 font encoding
\usepackage{cite}
\usepackage{amsmath,amssymb,amsfonts}
\usepackage{algorithmic}
\usepackage{graphicx}
\usepackage{textcomp}
\usepackage{xcolor}
\usepackage{booktabs}
\usepackage{array}
\usepackage[ruled]{algorithm2e}
\usepackage{subfigure}
\def\BibTeX{{\rm B\kern-.05em{\sc i\kern-.025em b}\kern-.08em
    T\kern-.1667em\lower.7ex\hbox{E}\kern-.125emX}}
\begin{document}

\title{Socially Adaptive Path Planning Based on Generative Adversarial Network}

\author{Yao Wang, Yuqi Kong, Wenzheng Chi*, and Lining Sun
\thanks{*corresponding author}

\thanks{This project is partially supported by National Key R$\&$D Program of China grant \#2020YFB1313601, National Science Foundation of China grant \#61903267 and China Postdoctoral Science Foundation grant \#2020M681691 awarded to Wenzheng Chi.}

\thanks{Yao Wang, Yuqi Kong, Wenzheng Chi, and Lining Sun are with the Robotics and Microsystems Center, School of Mechanical and Electric Engineering, Soochow University, Suzhou 215021, China {\tt\small \{ywang, yqkong, wzchi, lnsun\}@suda.edu.cn}}
}

\maketitle

\begin{abstract}
The natural interaction between robots and pedestrians in the process of autonomous navigation is crucial for the intelligent development of mobile robots, which requires robots to fully consider social rules and guarantee the psychological comfort of pedestrians. 
Among the research results in the field of robotic path planning, the learning-based socially adaptive algorithms have performed well in some specific human-robot interaction environments.
However, human-robot interaction scenarios are diverse and constantly changing in daily life, and the generalization of robot socially adaptive path planning remains to be further investigated. 
In order to address this issue, this work proposes a new socially adaptive path planning algorithm by combining the generative adversarial network (GAN) with the Optimal Rapidly-exploring Random Tree (RRT*) navigation algorithm.
Firstly, a GAN model with strong generalization performance is proposed to adapt the navigation algorithm to more scenarios.
Secondly, a GAN model based Optimal Rapidly-exploring Random Tree navigation algorithm (GAN-RRT*) is proposed to generate paths in human-robot interaction environments. 
Finally, we propose a socially adaptive path planning framework named GAN-RTIRL, which combines the GAN model with Rapidly-exploring random Trees Inverse Reinforcement Learning (RTIRL) to improve the homotopy rate between planned and demonstration paths. 
In the GAN-RTIRL framework, the GAN-RRT* path planner can update the GAN model from the demonstration path. 
In this way, the robot can generate more anthropomorphic paths in human-robot interaction environments and has stronger generalization in more complex environments.
Experimental results reveal that our proposed method can effectively improve the anthropomorphic degree of robot motion planning and the homotopy rate between planned and demonstration paths.

\end{abstract}
\def\abstractname{Note to Practitioners}
\begin{abstract}
The motivation of this work is to provide an efficient learning-based method to guide the robot generate the path in human-robot interaction environments.
Conventional path planning algorithms can quickly generate the optimal path from the start to the goal.
Although the robot did not collide with pedestrians, pedestrians are only regarded as moving obstacles, which greatly affects the psychological comfort of pedestrians.
In order to improve the anthropomorphic degree of robot path planning, we propose the GAN-RRT*.
It combines the RRT* path planning algorithm with the GAN model and learns anthropomorphic strategies from demonstration paths.
The proposed algorithm can be applied to service robot to improve the anthropomorphic degree.
\end{abstract}

\begin{figure}[t]
\centering
\includegraphics[width=3.0in]{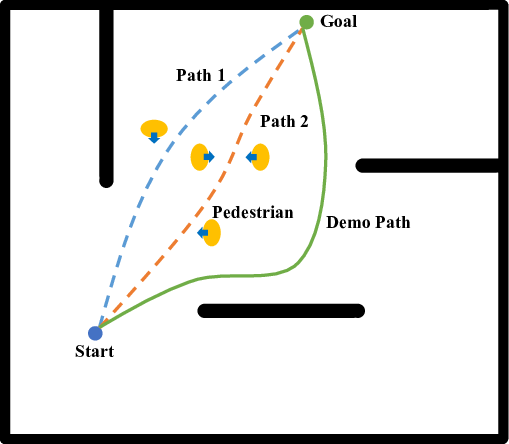}
\caption{A comparison of the demonstration path and two planned paths. The demonstration path is generated by the volunteer controlling the robot and the planned path are generated by RRT and A*, respectively. It is noteworthy that the demo path effectively avoid pedestrians comparing with other two paths.}
\label{fig1}
\end{figure}

\section{Introduction}
Over the last few decades, the autonomous navigation technology of mobile robots has made important breakthroughs. 
However, due to the lack of awareness of social rules, robots only regard pedestrians as movable obstacles in the navigation process. 
Some classic path planning algorithms such as the Dijkstra's\cite{D} algorithm, the A*\cite{A*} algorithm and the Rapidly exploring Random Tree (RRT) \cite{RRT} algorithm, can plan a short path to the target point, however, they still regard pedestrians as ordinary obstacles.
The navigation efficiency often decreases with the increase of pedestrian density in the environment. 
During the interaction, people still often need to compromise and accommodate robots, and their psychological comfort can not be guaranteed. 

As shown in Fig. \ref{fig1}, when planning a path in the human-robot interaction environment, people usually choose the demo path to avoid crowds. 
On the contrary, when only treating pedestrians as ordinary obstacles, the robot will generate path 1 or path 2 to shorten the path length. 
In order to endow mobile robots with the awareness of social rules and improve the degree of anthropomorphism of the robot path, many researchers initiated the study of robot social navigation\cite{4}\cite{5} and have been focused on generating a path similar to the demo path.
Currently, it is still difficult to comprehensively quantify and improve the similarity between the demo path and the generated path. 
Some methods may achieve good performance in certain scenario, however, the generalization and adaptability of navigation methods still need to be further improved in various human-robot interaction environments with high uncertainty.

%In complex human-robot interaction scenarios, the homotopy rate between the paths generated by the existing algorithms and the demonstration paths needs to be improved, and their generalization ability in complex scenarios needs to be strengthened.
Currently, the indicators for comparing two paths include the path length, the navigation time, the path dissimilarity\cite{14}, the feature count\cite{14}, the path cost\cite{14} and the Homotopy Rate\cite{homotopy_1}\cite{homotopy_2}.

\begin{definition}
(Homotopy Class of Trajectories):
Two trajectories with the same start and the same end are said to be in the same Homotopy Class if and only if one can be smoothly deformed into the other without intersecting obstacles. Otherwise, they belong to different homotopy classes.
\end{definition}

The Generative Adversarial Networks (GAN) model was originally proposed based on the game theory and showed good generalization ability in the field of image processing. 
Inspired by these methods, we present the GAN model based Optimal Rapidly-exploring Random Trees (GAN-RRT*) by combing the GAN model and the Optimal Rapidly-exploring Random Trees RRT*\cite{RRT*} algorithm. 
The GAN-RRT* path planner follows the general framework of the GAN model, which consists of two adversarial components: the generator and the discriminator. 
Through the inverse reinforcement learning process, the GAN-RRT* algorithm can generate more anthropomorphic paths and adapt to more scenarios by learning from the demonstration paths.

The rest of the paper is organized as follows: Section II introduces the problem of path planning and the algorithms of RRT and RRT*. Then Section III presents the details of the proposed GAN model, GAN-RRT* algorithm and GAN-RTIRL framework. The dataset collection and experimental results are reported in Section IV. We further discuss the results and the proposed system, and draw some conclusions at the end of this paper in Section V.

\subsection{Related Work}
In recent years, many scholars have proposed RRT-based algorithms to improve the efficiency of mobile robot path planning.
Chi \textit{et al.} \cite{1} proposed the Risk-DTRRT algorithm, which can optimize the global trajectory according to the trajectory prediction results of pedestrians in the environment.
Chi and Meng \cite{6} proposed the Risk-RRT* algorithm by combining the comfort and collision risk (CCR) map with the RRT* algorithm. 
The new Risk-RRT* method can achieve better results than Risk-RRT \cite{risk-rrt} in both static and dynamic environments.
Wang and Meng \cite{L-RRT*} proposed the L-RRT* algorithm, which can perform much better than the RRT* and Dynamic RRT* (D-RRT*) \cite{D-RRT*} in terms of time cost and path length.
The cost functions in the above RRT-based algorithms are pre-defined for specific scenarios, and only extract the distance from the pedestrian as a reference.
However, in the actual navigation process, the robot should regard the pedestrian as a motion model rather than a common moving obstacle.
%However, in the actual navigation process, the robot can not regard the pedestrian as an ordinary moving obstacle.

With the deepening of human-robot interaction, some scholars have focused on improving the comfort of pedestrians during the navigation process.
However, the rules of human society are usually behavioral tendencies, and often change according to different scenarios.
Therefore, it is crucial for the robot social-adaptive navigation to realize the digital expression of navigation behavior knowledge. %a big problem 
Inspired by Proxemics\cite{Proxemics}, Huang \textit{et al.} \cite{9} proposed a General Comfort Space (GCS) model with an elliptical shape, where the pedestrian is located at a focal point of the ellipse and the long and short axes of the ellipse are uniquely determined by the private distance and human speed in the spatial relations, respectively.
Chi \textit{et al.} \cite{10} proposed a Comfort and Collision Risk (CCR) map to unify the comfort of the pedestrians and the collision risk with the barriers in the environment into one risk map.
Truong and Ngo \cite{11} proposed a proactive social motion model (PSMM), which introduced social interaction space between crowds in robot motion planning to enable robots to navigate safely in dynamic crowded environments.
However, in human-robot interaction, it is not enough to consider only the pedestrian motion model.
Pedestrians cannot just be regarded as obstacles with a certain volume and moving speed. 
In the real world, pedestrians have their own minds and obey certain social rules.
Therefore, we still need to further study the rules of human-robot social interaction.

The movement rules of pedestrians will change to adapt to diverse human-robot interaction scenarios, which poses a great challenge to the quantification of the rules. 
In order to address this problem, some scholars introduce the end-to-end machine learning method \cite{GMR-RRT*,NeuralRRT*,hongyang} to the robot navigation framework.
Chen \textit{et al.} \cite{12} used deep reinforcement learning network to train robot motion strategies and solved three kinds of local obstacle avoidance problems (passing, crossing and overtaking) of robots.
Kim and Pineau \cite{13} proposed a local path planning method based on inverse reinforcement learning to plan motion paths according to different crowd densities and speeds.
Pérez-Higueras \textit{et al.} \cite{14,15,16,17} extracted the features of human-robot interaction for navigation tasks, and then obtained a more anthropomorphic robot path planning algorithm through inverse reinforcement learning framework.
Ding \textit{et al.} \cite{5} proposed a homotopy penalty strategy and combined it with inverse reinforcement learning to effectively reduce non-homotopic phenomenon during robot navigation.
In recent years, in order to improve the social adaptive ability of robots, many scholars combined generative adversarial network (GAN) to do further research on robot navigation methods.
Yang and Peters \cite{18} proposed a generative adversarial network structure App-GAN to generate a motion interaction mode for mobile robots when approaching a small group of pedestrians in conversation.
Gupta \textit{et al.} \cite{19} proposed a sequence-to-sequence social path generation method based on generative adversarial networks. The proposed Social-GAN enhanced the degree of anthropomorphism of paths in multi-person interaction environments.
Ma \textit{et al.} \cite{20} proposed conditional generative adversarial networks (CGAN) model to generate feasible paths to guide the sampling process of RRT*. 
However, the above algorithms focus on solving human-robot interaction problems in specific navigation scenarios. In real life, the scenarios and states of human-robot interaction are diverse and constantly changing, which requires our algorithm to have strong generalization performance. 
In addition, in order to further improve the comfort of pedestrians in the process of robot navigation, we also need to improve the homotopy rate between planned and demonstration paths.
%However, the navigation effect of the above methods in human-robot interaction environment needs to be improved. In particular, it is necessary to improve the homotopy rate between planned and demonstration paths and the generalization performance in complex scenarios.

\subsection{The Proposed System and Approach}
Inspired by the above methods, this paper proposes a new path planner with Generative Adversarial Networks based Optimal Rapidly-exploring Random Trees (GAN-RRT*). 
The growth of the RRT* tree depends on the tree node evaluation.
Traditionally, the node evaluation is usually carried out by the predefined cost function, such as the distance between two adjacent nodes and the cumulative distance from the tree root.
As aforementioned, given the complexity and the diversity of the human-robot interaction, it is difficult to comprehensively quantify the social-adaptive paths.
Therefore, we propose a generative network to calculate the cost value of growing nodes on the RRT* tree, which can make the cost function adaptable to more human-robot interaction scenarios.
%Our methodology is to generate paths by combining GAN and RRT* as a whole rather than separate the functions of the two parts. 
%Considering the process of inverse reinforcement learning, an adversarial network is used to judge the merits of each point along the path. 
Considering the process of inverse reinforcement learning, we need to give a criterion for the node cost value calculated by the generative network.
%A simple linear model cannot accurately judge whether the cost value of the path node is as expected.
In our previous work, we proposed to use the neural network to output the cost value and judge whether it is reasonable.
As aforementioned, due to the complexity of the training task, it is difficult for a single neural network to handle the inverse reinforcement learning task.
Therefore, we propose a adversarial network to judge whether the cost value of growing nodes on the RRT* tree is reasonable, which can offload the burden of the generative network and improve the performance of the generative network.
In order to formulate the social-adaptive path planning framework, we propose the GAN-RRT* based Inverse Reinforcement Learning (GAN-RTIRL) by combing the proposed generative and adversarial network. 
Through learning from the demo paths generated by volunteers remotely controlling the robot, the paths generated by GAN-RRT* will become more and more anthropomorphic with the GAN-RTIRL algorithm framework.
The strong generalization of the GAN model ensures that GAN-RRT* can be applied to more complex human-robot interaction environments. 
In the GAN-RTIRL training framework, GAN-RRT* can learn human-robot interaction rules from the demonstration path, thereby resulting in a higher homotopy rate between the planned path and the demonstration path.
Our work contributions can be summarized as follows:
\begin{itemize}
    \item A GAN model with strong generalization performance to adapt the navigation algorithm to more scenarios in human-robot interaction environments;
    \item A novel socially adaptive path planning algorithm named GAN-RRT* by combining the proposed GAN model and RRT*; and
    \item A GAN-RTIRL based social adaptive path planning framework for mobile robots to improve the degree of anthropomorphism.
\end{itemize}

\section{PRELIMINARIES}

\subsection{Path Planning Problem}
In this section, we elaborate the basic path planning problem.
Let $\mathcal X$ $\in$ $\mathbb{R}^n$ be the state space, $\mathcal X_{free}$ the free space and $\mathcal X_{obs}$ = $\mathcal X$  ${/}$  $\mathcal X_{free}$ the obstacle space. 
Let $x_{init}$ $\in$ $\mathcal X_{free}$ be initial state and $x_{goal}$ $\in$ $\mathcal X_{free}$ the goal state. 
In order to indicate whether the robot has reached the goal, we define the goal region as $\mathcal X_{goal}$ $\in$ $\mathcal X_{free}$. 
The path planning problem is to find a feasible path $\delta(t)$ $\in$ $\mathcal X_{free}$ for $t$ $\in$ [0,1], where $\delta(0)$ = $x_{init}$ and $\delta(1)$ $\in$ $\mathcal X_{goal}$.

%We also provide an explanation on how to obtain the optimal path. 
Let $\Sigma$ be the set of the feasible paths. 
To find the optimal path, we denote a cost function $c(\delta)$. 
The optimal path planning problem becomes finding the best path $\delta^*$ that minimizes the given cost function among $\Sigma$:
\begin{equation}
\label{eq:1}
\begin{aligned}
\delta^* = \mathop{\arg\min}\limits_{\delta \in \Sigma} c(\delta).\\
s.t. \ \delta(0) = x_{init}\\
\delta(1) \in \mathcal X_{goal}\\
\delta(t) \in \mathcal X_{free} 
\end{aligned}
\end{equation}

The process of finding the optimal path $\delta^*$ is actually the process of finding the set of optimal path nodes $N^*$. 
Finding the optimal set of path nodes $N^*$ is to find the path node $n^*$ with the lowest cost value.
In human-robot interaction environments, it is difficult to define an appropriate cost function for the RRT* algorithm. 
Therefore, in this article, we learn an end-to-end GAN model to calculate the cost of the path node instead.
\begin{equation}
\label{eq:2}
\begin{aligned}
c = cost(n) = GAN_g(n)\\
GAN(n) = GAN_d(n,c)\\
n^* = \mathop{\arg\max}\limits_{n \in \mathcal X_{free}} GAN(n)\\
\end{aligned}
\end{equation}

\begin{figure*}[t]
\centering
\includegraphics[width=6.5in]{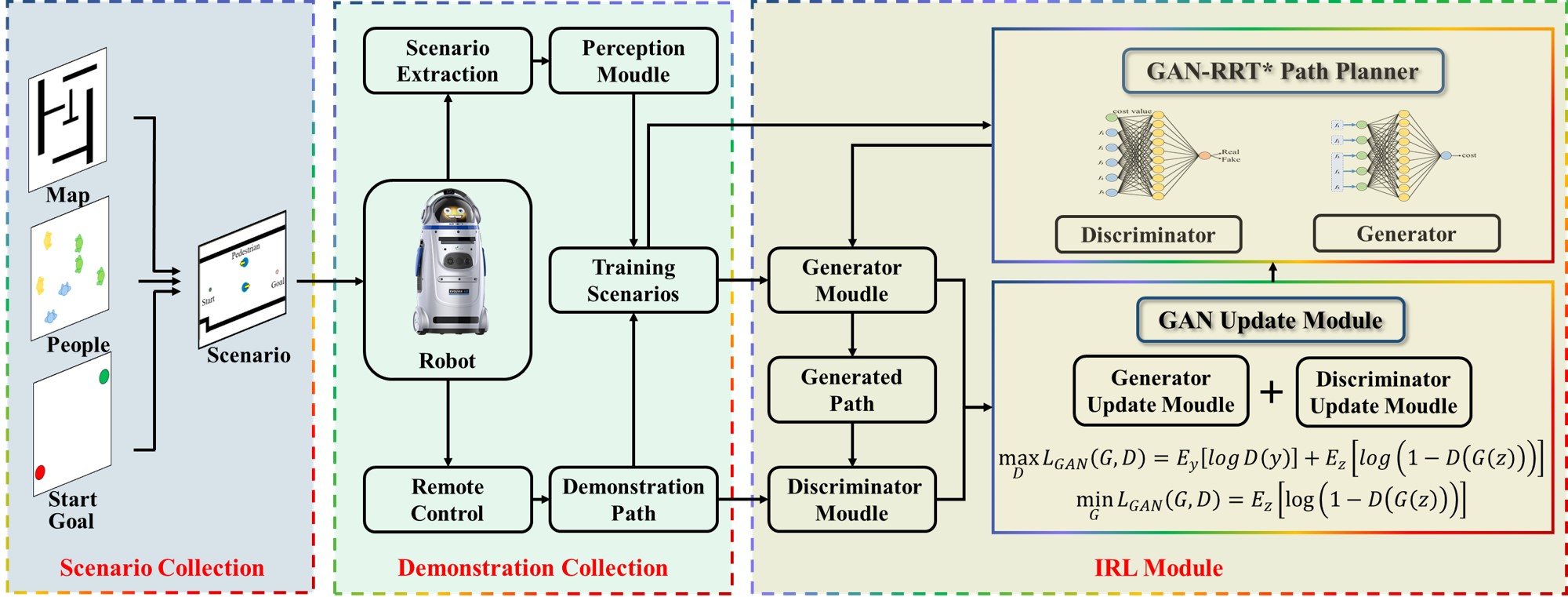}
\caption{Description of the proposed GAN-RTIRL framework.}
\label{fig5}
\end{figure*}

\subsection{The RRT* and NN-RRT* Algorithms}
The RRTs algorithm carried out path planning on a node tree, which is usually generated by randomly growing or biased sampling. 
The main process can be divided into four steps, namely random sampling, parent node selection, collision detection, and node connection. 
The specific process is explained as follows:

\textbullet\ Random sampling. Random sampling is carried out on the free area of the map to generate a new candidate node.

\textbullet\ Parent node selection. While traversing all existing nodes on the tree, the node closest to the candidate node is chosen as the parent node.

\textbullet\ Collision detection. The collision detection is conducted on the map between the growing node and the candidate node. If the line between two nodes encounters an obstacle, the collision detection failed.

\textbullet\ Connection. If the node completes collision detection, it will be added to the tree. Otherwise, the node will be dropped from the tree.

The RRT algorithm adopts Euclidean distance as the cost function and selects a node on the tree closest to the candidate node for connection. The RRT* algorithm adopts the cumulative Euclidean distance as the cost function. 
In other words, the process of selecting growth node not only considers the distance from the candidate node to the growing node, but also the cumulative distance from the root node to the growing node.

On the basis of the RRT algorithm, the RRT* algorithm adds two optimization steps: rewiring and re-connection.

\textbullet\ Rewiring. The new node will re-select its parent node to reduce its cumulative cost to the root node.

\textbullet\ Re-connection. If a new node is set as the parent node, we will change the parent of the node to the new node.

In the RRT algorithm, pedestrians are only considered as moving obstacles. 
The robot can only avoid collisions with pedestrians and ignore the psychological comfort of pedestrians.
%In the human-robot interaction environments, the robot cannot only consider the distance of the path, but also the comfort of pedestrians during navigation. 
In the RRT* algorithm, we proposed a cost function considering pedestrians and obstacles comprehensively in our previous work \cite{5}. 
The cost function considers elements related to social rules, including the distance to obstacles, the distance to pedestrians and the distance to the goal.
Considering that the linear model cannot accurately express the influence of the above factors on the interaction state of robots and pedestrians, 
%If we use the linear model to calculate the node cost value, the similarity between the planned path and the demonstration path is not high. 
we replaced the linear model with a fully connected neural network and proposed the new NN-RRT* \cite{22} algorithm, which greatly improves the similarity and homotopy rate between the planned path and the demonstration path. 
However, the generalization and adaptability of navigation methods still need to be further improved in various human-robot interaction environments with high uncertainty.
%Considering that the effect of these elements on the value of nodes may be non-linear, we replaced the linear model with neural networks and proposed the new NN-RRT* \cite{22} algorithm. 

\section{ALGORITHM}
In this section, we present the framework of GAN-RTIRL in Section III-A. 
Then we introduce the objective and structure of our GAN model in Section III-B. 
Finally, the details of GAN-RRT* path planner are shown in Section III-C.

\subsection{The GAN-RTIRL Framework}
In this work, we propose the GAN-RTIRL framework, as shown in Fig. \ref{fig5}. 
In the scenario collection phase, we integrate the information of map, people, start and goal into the scenario. 
Then, the volunteers control the robot to generate demonstration paths and the GAN-RRT* planner is utilized to generate planned paths. 
After that, we input the bag files of demonstration paths and generated paths into the training process. 
When the generator and discriminator in GAN-RRT* are in equilibrium, the features of the generated path and the demonstration path are increasingly similar and the path generator is output by IRL module.

In the IRL module, we pre-train the generator and discriminator to achieve a pre-balanced state.
The difference from previous work is that we no longer compare the global features of the two paths. 
Instead, we extract all node information of the two paths (Line 7-8 in algorithm 1). 
Firstly, we train the discriminator to accurately distinguish between the generated path nodes and the demonstration path nodes (Line 13 in algorithm 1). 
And then, the generator constantly update itself to confuse the discriminator and make it unable to distinguish between the generated path nodes and the demonstration path nodes (Line 14 in algorithm 1).
This process is repeated until the paths generated by the generator are similar to the demonstration paths before stopping the learning process.
Finally, a new GAN-RRT* planner is obtained, which can generate more anthropomorphic paths.
The complete pseudo code of the GAN-RTIRL is shown in Algorithm \ref{alg:2}.

\begin{algorithm}[t]
\LinesNumbered
\caption{GAN-RTIRL}
\label{alg:1}
\KwIn{Demo paths $\zeta_{plan}$ = $\left\{\zeta_{D1},...,\zeta_{DS}\right\}$ from S scenarios}
\KwOut{Generator $G$, Discriminator $D$}
    $G$ $\leftarrow$ Pre-training\\
    $D$ $\leftarrow$ Pre-training\\
    \While {$G$ {not converge}}
    {
        \For {s $\in$ S}
        {
            \For {repetitions}
            {
              $\zeta_{plan}$ $\leftarrow$ GAN-RRT*($s$)\\
              $N_{plan}$ $\leftarrow$ GetNode($\zeta_{plan}$)\\
              $N_{demo}$ $\leftarrow$ GetNode($\zeta_{demo}$)\\

            }
            $\sigma_{plan}$ $\leftarrow$ $\sigma$ $\cup$ $N_{plan}$\\
            $\sigma_{demo}$ $\leftarrow$ $\sigma$ $\cup$ $N_{demo}$
        }
            $D$ $\leftarrow$ Update-Discriminator($\sigma_{plan}$, $\sigma_{demo}$)\\
            $G$ $\leftarrow$ Update-Generator($\sigma_{plan}$)\\
    }   
    \Return $G$ and $D$
\end{algorithm}

\subsection{The GAN-based Socially Aware Path Node Evaluation}
%Although a simple fully connected neural network can guide RRT* path planning and work well in human-robot interaction environments with fewer pedestrians. 
%However, it is not satisfactory in complex scenes because of the weak generalization performance of a single neural network. 
%In addition, there are still differences between the path planned by NN-RRT* and the demonstration path. 
%The generalization performance of NN-RRT* in complex environments needs to be improved.
The socially aware RRTs planner works well in human-robot interaction environments with a few pedestrians. 
However, their generalization performance in complex environments needs to be further improved.
In this work, we propose a GAN-RRT* planner to overcome these issues.
The GAN model has strong generalization performance when dealing with complex problems, which consists of two neural networks that compete against each other. 
The two networks are trained cooperatively through competitive learning to improve the performance of the generative network.
The generator is used to deceive the discriminator, while the discriminator is used to judge the ability of the generator. 
Finally, the ability of the two models becomes stronger and reaches a steady state.

The GAN model is originally designed to learn a mapping from random noise $z$ to the output image $y$.
In the GAN-RRT* algorithm, the GAN model is designed to learn the mapping from feature vector $f$ to the output node cost value $c$.
The generator $G$ is defined as \{$G$ : $f$ $\rightarrow$ $c$ \}. 
The value of $c$ is calculated by $GAN_g(f)$ in Eq. \ref{eq:2}.
The discriminator $D$ is designed to distinguishing whether the node cost value generated by $G$ is real or fake. 
The probability computed by $D$ that $c$ is the real data is defined as $D(c)$ corresponding to $GAN_d(n,c)$ in Eq. \ref{eq:2}.
The objective function can be expressed as:
\begin{equation}
\label{eq:3}
\begin{aligned}
\mathcal{L}_{GAN}(G,D)= \mathbb{E}_c[logD(c)]+\mathbb{E}_f[log(1-D(G(f)))].
\end{aligned}
\end{equation}

The GAN aims to train $D$ to maximize the log$D(c)$ and train $G$ to minimize log$(1-D(G(f)))$. The optimization function can be expressed as:
\begin{equation}
\label{eq:4}
\begin{aligned}
\theta^*=arg \mathop{min}\limits_{G} \mathop{max}\limits_{D} \mathcal{L}_{GAN}(G,D).
\end{aligned}
\end{equation}

It is noteworthy that the generator $G$ and discriminator $D$ can be various neural networks, without any limitation on the both models of adversarial networks. Therefore, we can modify the network structure of the GAN model to adapt our algorithm.

%\subsection{Architectures of Generator and Discriminator}
%The GAN-RRT* will select nodes with lower cost value for growth. 
When the robot generates path nodes, the GAN-RRT* algorithm calculates the cost value of the current node through the $GAN_g$ function according to the environment information around the node. 
If the path node is away from obstacles and pedestrians and towards the goal, the result of the output of the function $GAN_d$ is closer to 1.
In general, the GAN module can screen out socially aware path nodes.

\begin{figure}[t]
\centering
\includegraphics[width=3.0in]{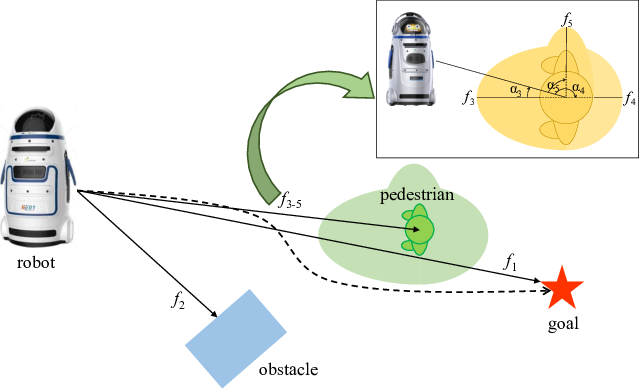}
\caption{The features of interaction between the robot and the pedestrian.}
\label{fig2}
\end{figure}

To represent the interaction state between the robot and the environment in the human-robot interaction, we adopt a set of socially aware features, as shown in Fig. \ref{fig2}.
\begin{itemize}
\item $f_1$ represents the distance from the robot to the goal.
\item $f_2$ represents the distance from the robot to the nearest obstacle.
\item $f_{3-5}$ represent the cost related to the pedestrian. The cost function is defined by the Gaussian function, which is divided into three directions: front, back and right of the pedestrian, respectively.
\end{itemize}

\begin{figure}[t]
\centering
\includegraphics[width=3in]{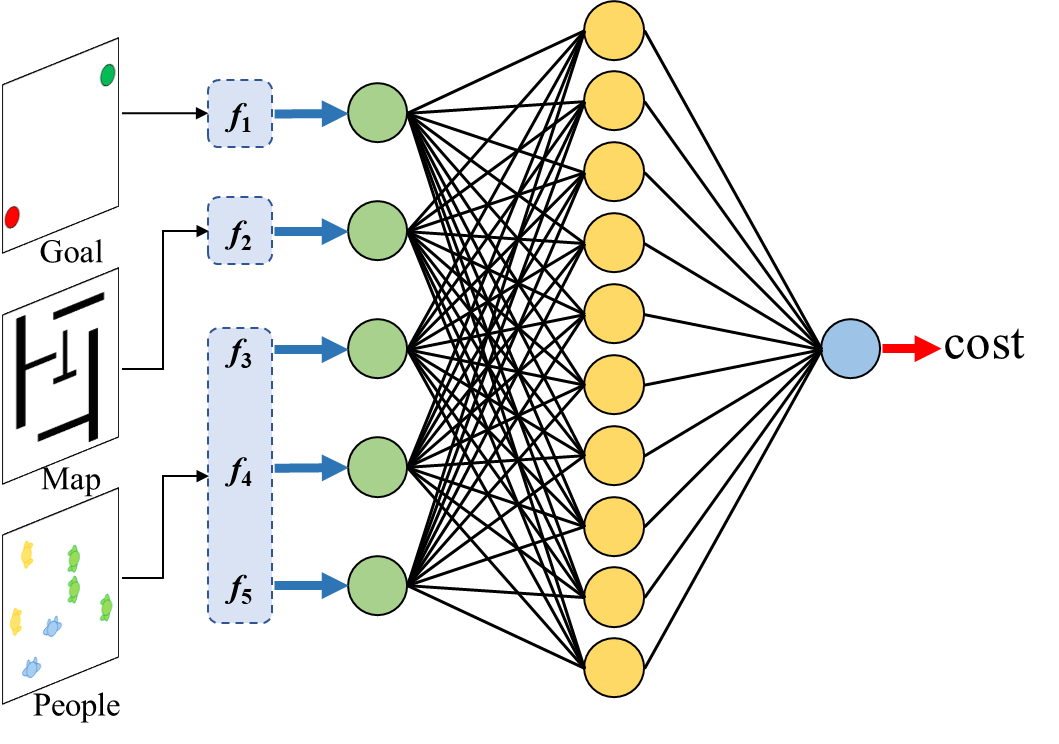}
\caption{The architecture of the generator. The input layer is the feature information of the map, goal and people detected by the robot. The output layer is the cost value of the current node of the robot.}
\label{fig3}
\end{figure}

\begin{figure}[t]
\centering
\includegraphics[width=3.2in]{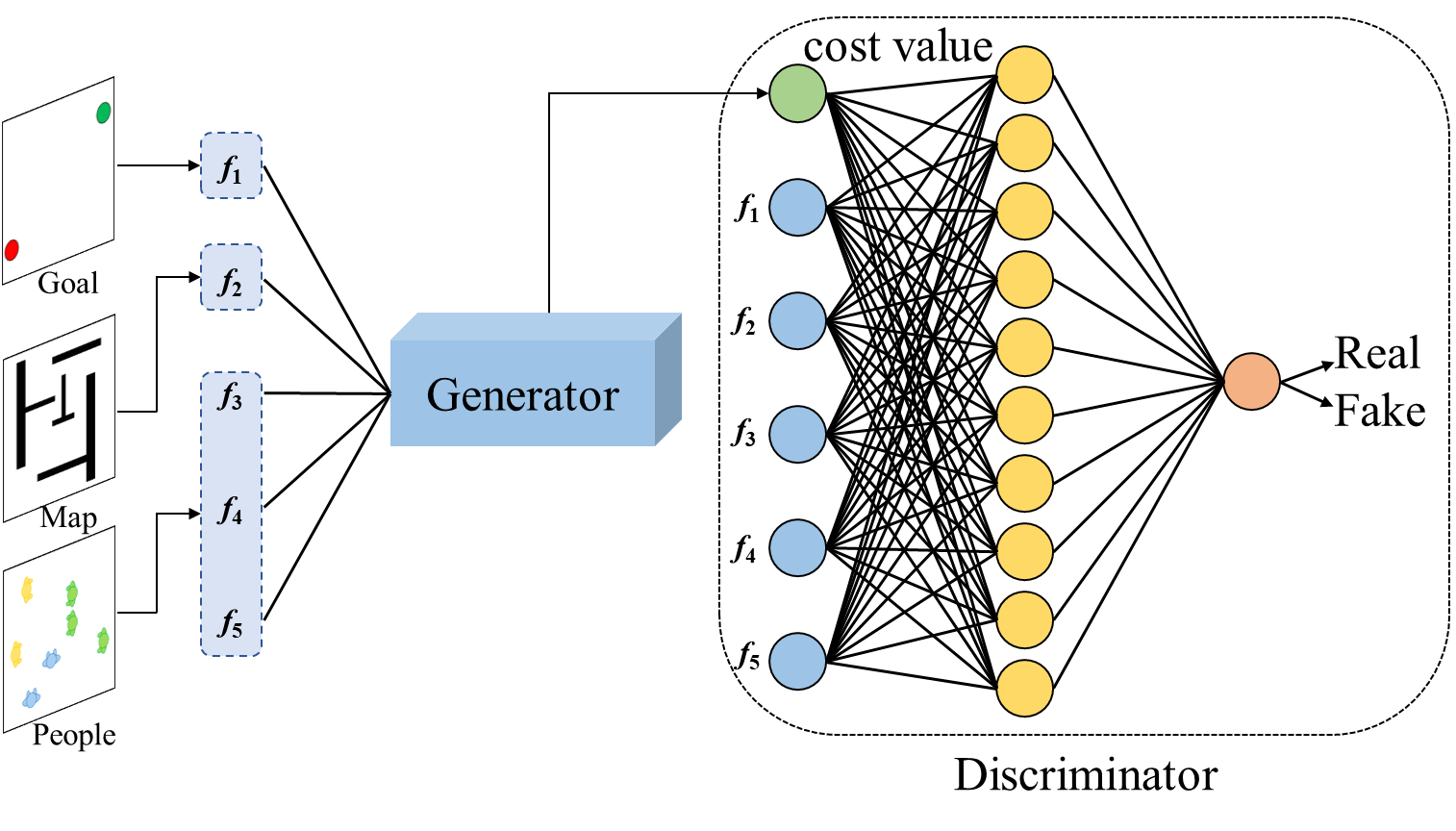}
\caption{The architecture of the discriminator. The input layer is the feature information and cost value of the current node. The output layer is the discriminant result of the current node.}
\label{fig4}
\end{figure}

The input of the generator in GAN-RRT* includes the five features mentioned above, which aims to calculate the cost value of the RRT* node in the human-robot environment accurately. 
As shown in Fig. \ref{fig3}, there are five neurons corresponding to $f_1$, $f_2$, $f_3$, $f_4$, $f_5$ in the input layer and the the output of the generator is the cost value of the node. 
We add ten neurons to the hidden layer where the activation function is Sigmod. 
%The mathematical expression of the Sigmoid activation function is $f(x)=\frac{1}{1+e^{-x}}$, where $x$ is the input value. The architecture of generator is .
The architecture of discriminator is shown in Fig. \ref{fig4}. 
Similar to the generator, there are six neurons in the input layer of discriminator and the the output of the discriminator is the probability that the cost value of the node is real.
With the information of map, pedestrians and goal, the generator can get the cost value of the node. 
And then, the discriminator will calculate these five features and the cost value to obtain a probability value. 
The function of the discriminator is to distinguish whether the cost value derived from the generator is ideal. 
During the training process, the abilities of the two models become increasingly stronger, ultimately constraining each other to reach a balanced state.

\begin{algorithm}[t]
\LinesNumbered
\caption{The GAN-RRT* Algorithm}
\label{alg:2}
\KwIn{Start, Goal, Map and GAN Model $G$}
\KwOut{The GAN-RRT* $Tree = (V, E)$}
    $V$ $\leftarrow$ $\left\{x_{init}\right\}$; $E$ $\leftarrow$ $\emptyset$;\\
    \For {i = {\rm 1},...,n}
        {
        $x_{rand}\leftarrow Sample_i$;\\
        $x_{nearest}\leftarrow$ Nearest($Tree$, $x_{rand}$);\\
        $x_{new} \leftarrow$ Steer($x_{nearest}$, $x_{rand}$);\\
        \If{ObstcalFree($x_{new}$)}
            {
             $X_{near}\leftarrow$ Near($Tree$, $x_{new}$, $r$);\\
             $V \leftarrow V \cup \left\{x_{new}\right\}$;\\
             $x_{min} \leftarrow$ $x_{nearest}$;\\
             $c_{min} \leftarrow$ $C_{G} (x_{nearest}, x_{new})$ ;\\
             \If{CollisionFree($x_{new}, x_{near}$)}
                {
                \For {$x_{near}$ $\in$ $X_{near}$}
                    {
                    \If{$ {\rm C_{G}}(x_{near},x_{new}) < c_{min}$}
                        {
                        $x_{min}$ $\leftarrow$ $x_{near}$;\\ 
                        $c_{min}$ $\leftarrow$ 
                        ${\rm C_{G}}(x_{near},x_{new})$;
                        }
                    }
                $E$ $\leftarrow$ $E$ $\cup$ $\left\{(x_{min}, x_{new})\right\}$;\\
                \For {$x_{near}$ $\in$ $X_{near}$}
                    {
                    \If{${\rm C_{G}}(x_{new},x_{near}) < {\rm Cost_{G}}(x_{near})$}
                        {
                        $x_{parent} \leftarrow SetParent(x_{near})$;\\
                        $E \leftarrow (E \backslash \left\{(x_{parent}, x_{near})\right\}) \cup \left\{(x_{new}, x_{near})\right\}$;
                        
                        }
                    }
                }    

            }
        }
\Return $Tree$
\end{algorithm}

\subsection{GAN-RRT* Path Planner}
The trained generator model is used in the sampling process of GAN-RRT* algorithm. 
The Algorithm 2 shows the details of the GAN-RRT* algorithm.
Taking the start point as the root node, the algorithm gets $x_{rand}$ through $Sample$ function.
We can find the nearest node to $x_{rand}$ through the $Nearest$ function and get $x_{new}$ through the $Steer$ function. 
Then, we identify candidate nodes within a certain range $r$.
After the introduction of generative adversarial network, GAN-RRT* algorithm can calculate the cost value of the current node through the $GAN_g$ function and distinguish whether the current cost value is consistent with the current state through the $GAN_d$ function. 
In addition, the GAN-RRT* algorithm can also calculate the cumulative cost value of the current node based on candidate nodes.
Through the above process, the new node $x_{new}$ can re-select its parent node $x_{parent}$ to optimize the current generated path. 
After re-selecting the new parent node, the node $x_{new}$ also can choose its growing node to reduce the cumulative cost value of the growth node. 
In the process, the candidate nodes can select $x_{new}$ as their parent node corresponding to lines 19 to 24 in algorithm 2.

As the sampling process repeats, the cost value of the obtained path tree decreases and the GAN-RRT* algorithm will eventually generate an optimal path.

\begin{figure}[t]
\centering
\includegraphics[width=3in]{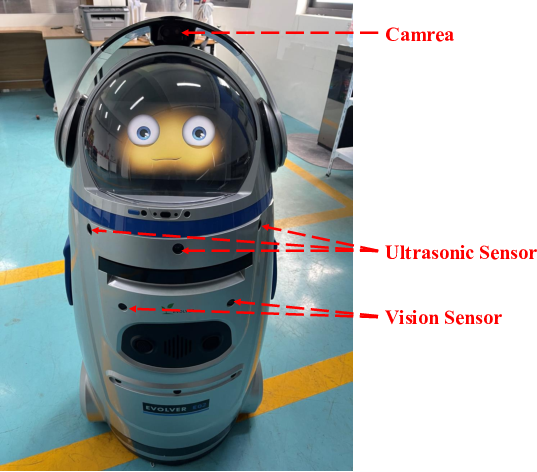}
\caption{Xiaopang robot equipped with a depth camera in the real-world experiments.}
\label{fig9}
\end{figure}

\section{EXPERIMENTAL STUDIES AND RESULTS}
In this work, we carry out simulation and real-world experiments based on the Robot Operating System (ROS)\cite{ROS} with Ubuntu 16.04 on a laptop equipped with Intel i7-11800H CPU @2.3 GHz and 16 GB of RAM. 
Firstly, we design static maps and produce demonstration paths in the simulation environment. 
Secondly, we demonstrate the performance of the GAN-RRT* through several numerical simulations. 
Finally, we evaluate the planning performance of the proposed algorithm in 100 human-robot interaction environments.
In the real-world experimental studies, a robot named Xiaopang is adopted, as shown in Fig. \ref{fig9}.

To better evaluate the path generated by GAN-RRT*, we select three metrics for objective evaluation: the dissimilarity, the feature difference and the homotopic rate.
The dissimilarity is numerically equal to the area of a closed region enclosed by two paths. 
\begin{equation}
\begin{aligned}
dissi&milarity({\zeta}_1,{\zeta}_2) = \\
&\frac{\sum_{k=1}^{N-1}\frac{d({\zeta}_1(k),{\zeta}_2)+d({\zeta}_1(k+1),{\zeta}_2) \left|\left|{\zeta}_1(k)-{\zeta}_1(k+1)\right|\right|}{2}}{\delta}         
\label{diss}
\end{aligned}
\end{equation}
The $d$ function is used to calculate the distance between two path nodes. $\delta$ is the number of segments of the path ${\zeta}_1$.

The feature difference is numerically equal to the absolute value of the feature difference between the two paths. 
\begin{equation}
F_{fd} = \frac{1}{5*S}\sum_{i=1}^S\sum_{j=1}^{5}\left|f^{demo}_{(i,j)}-f^{plan}_{(i,j)}\right|         \label{fc}
\end{equation}
$S$ represents the number of the testing scenarios.

The homotopy rate of generated paths and demonstration paths is defined as the proportion of generated paths within the same homotopy class of the demonstration path.

%For the diversity of experimental scenarios, we design 100 different static maps. 
%The size of each map is 324 pixel × 257 pixel. 
%The start, goal, and pedestrian positions are designed according to the situations of human-robot interaction in the real environments. Finally, the scenarios are visualized in RVIZ, and then the volunteers control the robot to generate the demonstration paths.

%For the diversity of experimental scenarios, we design 100 different static maps. 
%The size of each map is 324 pixel × 257 pixel. 
%The start, goal, and pedestrian positions are designed according to the situations of human-robot interaction in the real environments. Finally, the scenarios are visualized in RVIZ, and then the volunteers control the robot to generate the demonstration paths.
\subsection{Performance in Simulations}
In this section, we design 100 different static maps. The size of each map is 324 pixel × 257 pixel.
And then, we compare the navigation effects of the RRT*, the NN-RRT* and the GAN-RRT* in simulations. 
Among the 100 scenarios, we selected 75 scenarios as the training set and 25 scenarios as the testing set. 
On the testing set, the homotopic rate of RRT* is 75$\%$, the rate of NN-RRT* is 90$\%$ and the rate of GAN-RRT* is 94$\%$. The experimental results show that compared with RRT* and NN-RRT* methods, GAN-RRT* method achieves higher homotopy rate and higher anthropomorphic degree of generated paths.

\begin{figure}[t]
\centering
\subfigure[Difference of dissimilarity.] {
\label{fig:back1}
\includegraphics[width=2.5in]{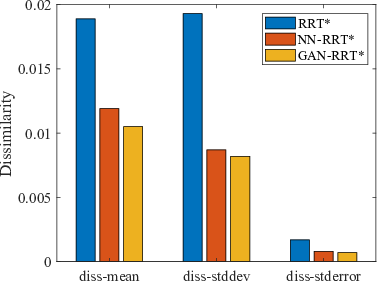}
}
~
\subfigure[Difference of features.] {
\label{fig:back2}
\includegraphics[width=2.5in]{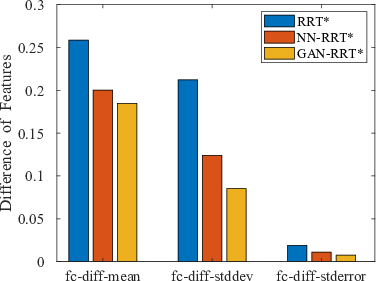}
}
~
\caption{Difference of dissimilarity and features among RRT*, NN-RRT* and GAN-RRT*.}
\label{fig7}
\end{figure}

As shown in Fig. \ref{fig7}(a), in terms of dissimilarity, the dissimilarity of GAN-RRT* is the lowest among the three methods, which means that the area of a closed region enclosed by the path generated by GAN-RRT* and the demonstration path is the smallest.
As shown in Fig. \ref{fig7}(b), in terms of feature difference, the feature difference of GAN-RRT* is also the lowest among the three methods, which means that the paths generated by the GAN-RRT* planner are more similar to demonstration
paths.
With the increase of the number of testing scenarios, the paths generated by GAN-RRT* planner still maintain a high degree of anthropomorphism thanks to the strong generalization performance of generative adversarial networks.

\begin{figure*}[t]
\centering
\includegraphics[width=6.2in]{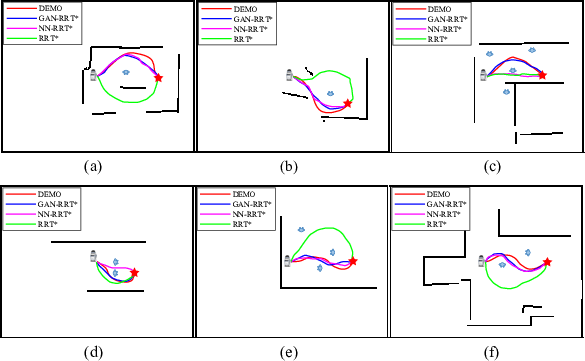}
\caption{Comparison of paths generated by RRT*, NN-RRT*, GAN-RRT* and volunteers.}
\label{fig8}
\end{figure*}

\begin{figure*}[t]
\centering
\includegraphics[width=6.2in]{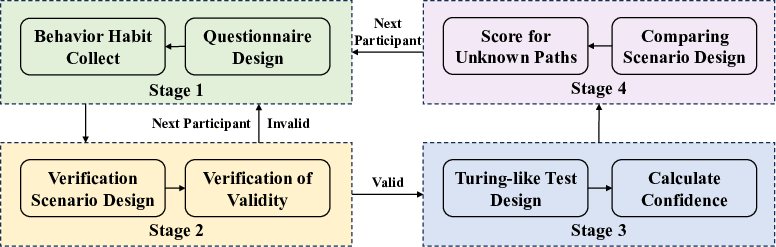}
\caption{The general scheme of the questionnaire.}
\label{fig10}
\end{figure*}

Fig. \ref{fig8} provides a more intuitive representation of the difference between the planned paths and the demonstration paths.
As shown in Fig. \ref{fig8}(a) and Fig. \ref{fig8}(b), the paths generated by algorithms of GAN-RRT* and NN-RRT* can pass behind pedestrians and reach the goal in uncomplicated human-robot interaction scenarios. 
This follows the behavioral norms that avoid disturbing pedestrians in daily life. 
However, pedestrians often appear in the form of crowds, which requires the robot to plan the path as far as possible to avoid the people who are communicating. 
As shown in Fig. \ref{fig8}(c)-(f), the algorithm GAN-RRT* also performs well in complex crowd environments. 

\begin{figure*}[t]
\centering
\includegraphics[width=6.2in]{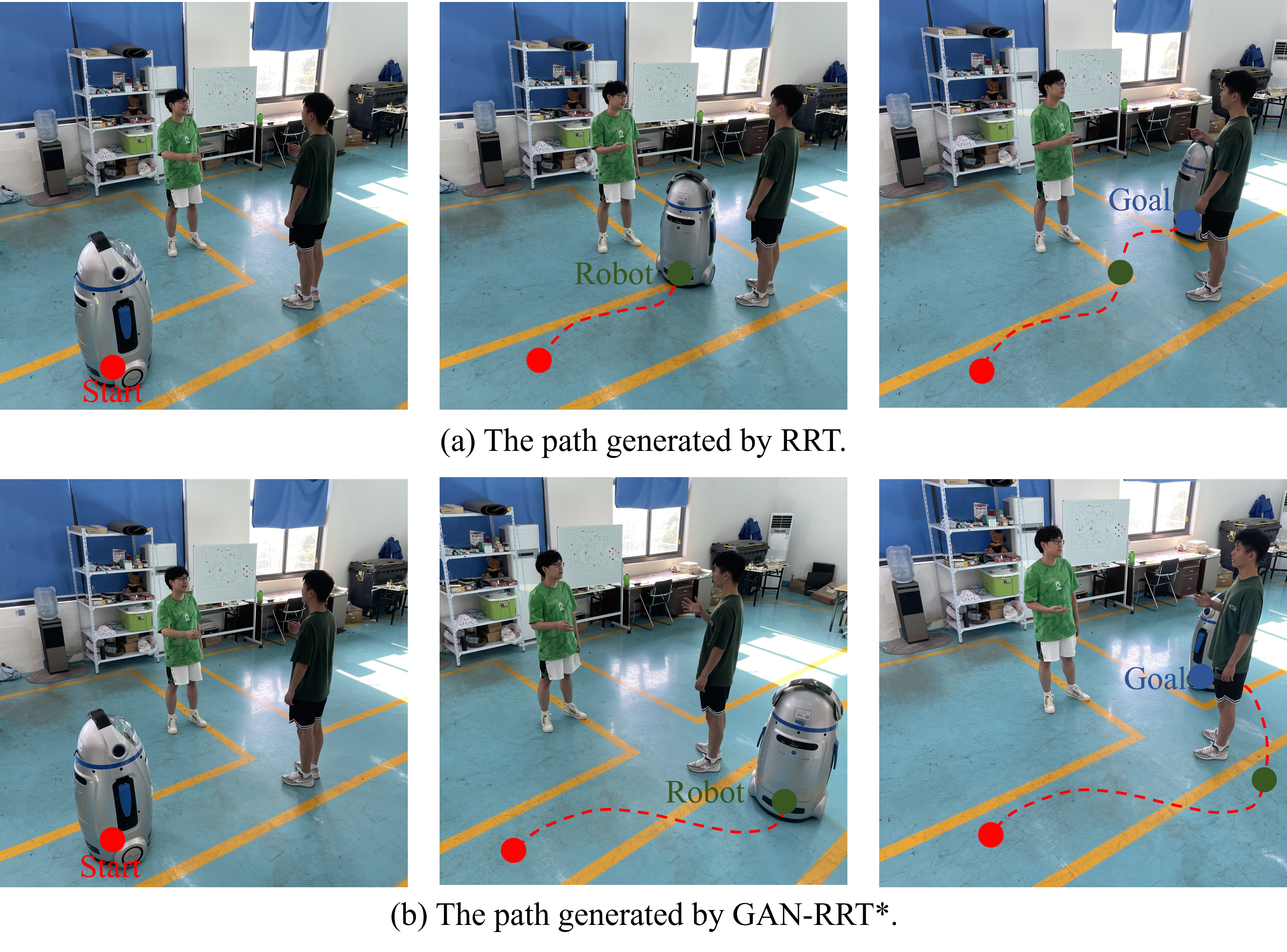}
\caption{Turing-like test: Estimate the anthropomorphic confidence of path generated by GAN-RRT*.}
\label{fig11}
\end{figure*}

\subsection{Performance in Real-World Experimental Studies}
In the real-world experimental studies, we adopt the investigation questionnaire to evaluate the socially aware performance of the propose method.
As shown in Fig. \ref{fig10}, the investigation questionnaire is divided into four stages. 
We invited 30 students from different majors to complete the survey. 
Firstly, we propose a series of issues related to navigation in human-robot interaction environments. 
Secondly, participants choose one of the two paths to verify the validity of their results in the first questionnaire survey. 
Thirdly, we investigate the anthropomorphic confidence of paths generated based on the GAN-RRT* in the third stage.
Finally, we utilize a Likert scale to compare paths generated by GAN-RRT*, NN-RRT* and RRT*.

 From the perspective of psychology, people tend to score higher on their preferred paths and believe that these paths have a high degree of anthropomorphism.
Therefore, we further verify the superiority of our navigation algorithm by combining questionnaires and real-world experiments.

\textbf{Stage 1-Collect Behavior habits}

In order to collect the behavioral habits of participants in human-robot interaction environments, we designed three questions as shown in Appendix.
From the relative results, it can be analyzed that 80$\%$ of people believe that robots need to pay more attention to human-robot interaction status when navigating. 
In complex environments, 86.67$\%$ of people choose to bypass the crowd, and 13.33$\%$ choose to pass through the crowd. 
In addition, it is generally believed that robots need to consider more factors during the navigation, such as the distance between the robot and the pedestrian, the position of the robot relative to the human, the velocity of the robot relative to humans and the relationship between the path and human state. 
However, paper-based questionnaires do not accurately reflect the real thoughts of participants. Only when participants see the real planning path can they give a relatively objective  judgment.
Therefore, we designed the second stage to verify the validity of the survey questionnaire.

\textbf{Stage 2-Questionnaire Validation}

Because people can always choose their preferred path in comparison, we present both the demonstration path and the RRT-generated path in the same scenario. 
Participants will choose the path they prefer between the two paths.  
If participants choose to pass through the crowd in the first stage and choose the path generated by RRT in the second stage, then the questionnaire survey is considered valid. 
Similarly, participants choose to cross from a sparsely populated area and choose the demonstration path, the questionnaire survey is also valid. 
Otherwise, we believe that the questionnaire survey is invalid.

\textbf{Stage 3-The Anthropomorphic Confidence of Paths Generated by GAN-RRT* Algorithm}

Inspired by the Turing test\cite{turing}, we propose a anthropomorphic confidence calculating method for hybrid testing of multiple algorithms.
As shown in Fig. \ref{fig11}, participants do not know the algorithm used for robot navigation beforehand. 
For example, the path through the crowd in Fig. \ref{fig11}(a) is generated by RRT and the path bypass the crowd in Fig. \ref{fig11}(b) is generated by GAN-RRT*. 
In the same scenario, the robot travels in the real environment along the demonstration path and the path generated by RRT and GAN-RRT*, respectively. 
After each trial, the participants are required to choose the demonstration path from the three paths. 
If the participant believe that the path generated by GAN-RRT* is a demonstration path,
this means that the GAN-RRT* generated path is misjudged as the demonstration path.

We provide the following definition for anthropomorphic confidence of paths generated by GAN-RRT* algorithm:
\begin{equation}
F_{ac} = \sum_{i=1}^S\sum_{j=1}^{P_s}\frac{mis(i,j)}{n_G(i,j)*P_S}         \label{eq4}
\end{equation}

\begin{equation}
mis(i,j)=
\begin{cases}
1,& \text{ $ Path \ misjudged $ } \\
0,& \text{ $ Path \ not \ misjudged $ }
\end{cases}
\label{eq5}
\end{equation}
Herein, $S$ represents the number of scenarios, $P_S$ represents the number of participants in this scenario, $mis(i,j)$ indicates that the path generated by the GAN-RRT* is misjudged as the demonstration path and $n_G(i,j)$ represents whether the path is generated by the GAN-RRT*.

In this stage, participants are required to identify demonstration paths and robot paths in total 10 scenarios. 
The final anthropomorphic confidence of social navigation scheme based on GAN-RRT* is 80.77$\%$, which also proves that our proposed navigation algorithm can achieve good results in human-robot interaction environments.

\textbf{Stage 4-Comparision between Paths based on RRT* and NN-RRT* and GAN-RRT*}

In order to intuitively shown the performance of the navigation methods, participants are further required to rate the paths generated by RRT*, NN-RRT* and GAN-RRT*.
Since it is difficult for participants to accurately describe their acceptance of each path with words, we adopt Likert scale to give scores for each path.
Participants can score for the path between 1 and 5 according to their behavior habits based on the Likert scale. 

\begin{figure}[t]
\centering
\includegraphics[width=2.8in]{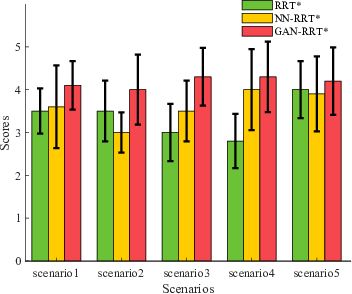}
\caption{The average scores of RRT*, NN-RRT* and GAN-RRT* in all 5 scenarios.}
\label{fig12}
\end{figure}

We selected 20 participants to conduct comparative experiments in 5 scenarios. 
After the robot generates three paths in one scenario, participants are asked to score for each path. 
As shown in Table \ref{tab3}, the average scores of RRT*, NN-RRT* and GAN-RRT* are 3.36, 3.6 and 4.18, respectively. 
The scores of RRT*, NN-RRT* and GAN-RRT* in 5 scenarios are shown in Fig. \ref{fig12}. 
From the experimental results, it can be seen that both the average scores and the scores in 5 scenarios, GAN-RRT* is the highest. 
This also indicates that the path generated by GAN-RRT* makes pedestrians more comfortable in human-robot interaction environments.

\begin{table}[t]
\caption{The scores of RRT*, NN-RRT* and GAN-RRT*.}
\setlength{\tabcolsep}{2pt}
\centering
\begin{tabular}{cccccc}
    \toprule
    Environment & Map Size & Pedestrians & RRT* & NN-RRT* & GAN-RRT* \\
    \midrule
    Scenario1 & 5m $\times$ 5m & 3 & 3.5  & 3.6  &4.1   \\
    Scenario2 & 5m $\times$ 5m & 3 & 3.5  & 3    &4   \\
    Scenario3 & 5m $\times$ 5m & 3 & 3    & 3.5  &4.3   \\
    Scenario4 & 5m $\times$ 5m & 3 & 2.8  & 4    &4.3   \\
    Scenario5 & 5m $\times$ 5m & 3 & 4    & 3.9  &4.2   \\
    Average   & 5m $\times$ 5m & 3 & 3.36 & 3.6  &4.18   \\
   \bottomrule
\end{tabular}
\label{tab3}
\end{table}

\section{CONCLUSION}
This article has proposed a sampling-based path planning algorithm GAN-RRT* to guide robot to navigate in human-robot interaction environments. 
Firstly, a generative adversarial network model with strong generalization performance is proposed to adapt the navigation algorithm to more scenarios. 
Secondly, we propose the new navigation algorithm GAN-RRT* by combining the GAN model with RRT* to guide robots to generate paths in human-robot interaction scenarios.
Finally, the socially adaptive path planning framework GAN-RTIRL is proposed to improve the generalization performance of robot navigation and the homotopy rate between the generated path and the demonstration path.
We conducted comparative experiments in simulation and real world.
The experimental results have revealed that the proposed navigation algorithm greatly improves the generalization performance of robot navigation in human-robot interaction environments and the homotopy rate between the generated path and the demonstration path.
In complex and densely populated environments, the paths generated by our method are more socially aware and anthropomorphic.

\appendix

\begin{table}[ht]
\centering
\resizebox{\linewidth}{!}{
\begin{tabular}{l}
    \toprule
    \centerline{Questionnaire}  \\
    \midrule
    Question 1: \\Which factors need to be considered more when robots navigate\\
    in human-robot  interaction environments? \\
    Answer 1:\\
    (a)Path length; \\
    (b)Human-robot interaction state. \\ 
    \\

    Question 2: \\Which navigation strategy would you choose when robots\\
    navigate in a complex human-robot interaction environment? \\
    Answer 2:\\
    (a)Keep the path length as short as possible; \\
    (b)Pass through sparsely populated areas.  \\
    \\

    Question 3: \\Which of the following factors should be considered in the\\
    process of human-robot interaction?   \\
    Answer 3:\\
    (a)The distance between the robot and the person; \\
    (b)The position of the robot relative to the human; \\
    (c)The movement speed of robots relative to humans; \\
    (d)The relationship between the path and human state. \\
    \\
    Question 4: \\
    The robot will generate paths twice based on different \\
    algorithms. Please choose the path you prefer from the next \\
    two paths and write down your answer.\\
    Answer 4:\\
    (a)The first path.\\
    (b)The second path.\\
    Note: Please give your choice after both paths are generated.  \\
    \\
    Question 5: \\
    The robot will generate paths three times based on different \\
    algorithms. Please choose the path that you think is artificially\\
    generated from the next three paths and write down your answer.\\
    Answer 5:\\
    (a)The first path.\\
    (b)The second path.\\
    (c)The third path.\\
    Note: Please give your choice each time the path is generated.  \\
    \\
    Question 6: \\
    We will let the robot generate paths three times.\\
    Please express how much you agree with the generated path. \\
    One point represents strong disagreement, two points represent \\
    disagreement, three points represent common, four points \\
    represent agreement and five points represent strong agreement.\\
    Answer 6:\\
    (a)Strong disagreement.\\
    (b)Disagreement.\\
    (c)Common.\\
    (d)Agreement.\\
    (e)Strong agreement.\\
    Note: Please give your choice each time the path is generated. \\
   \bottomrule
\end{tabular}}
\label{tab1}
\end{table}

\bibliographystyle{IEEEtran}      
\bibliography{wang}

\begin{IEEEbiography}
	[{\includegraphics[width=1in,height=1.25in,clip,keepaspectratio]{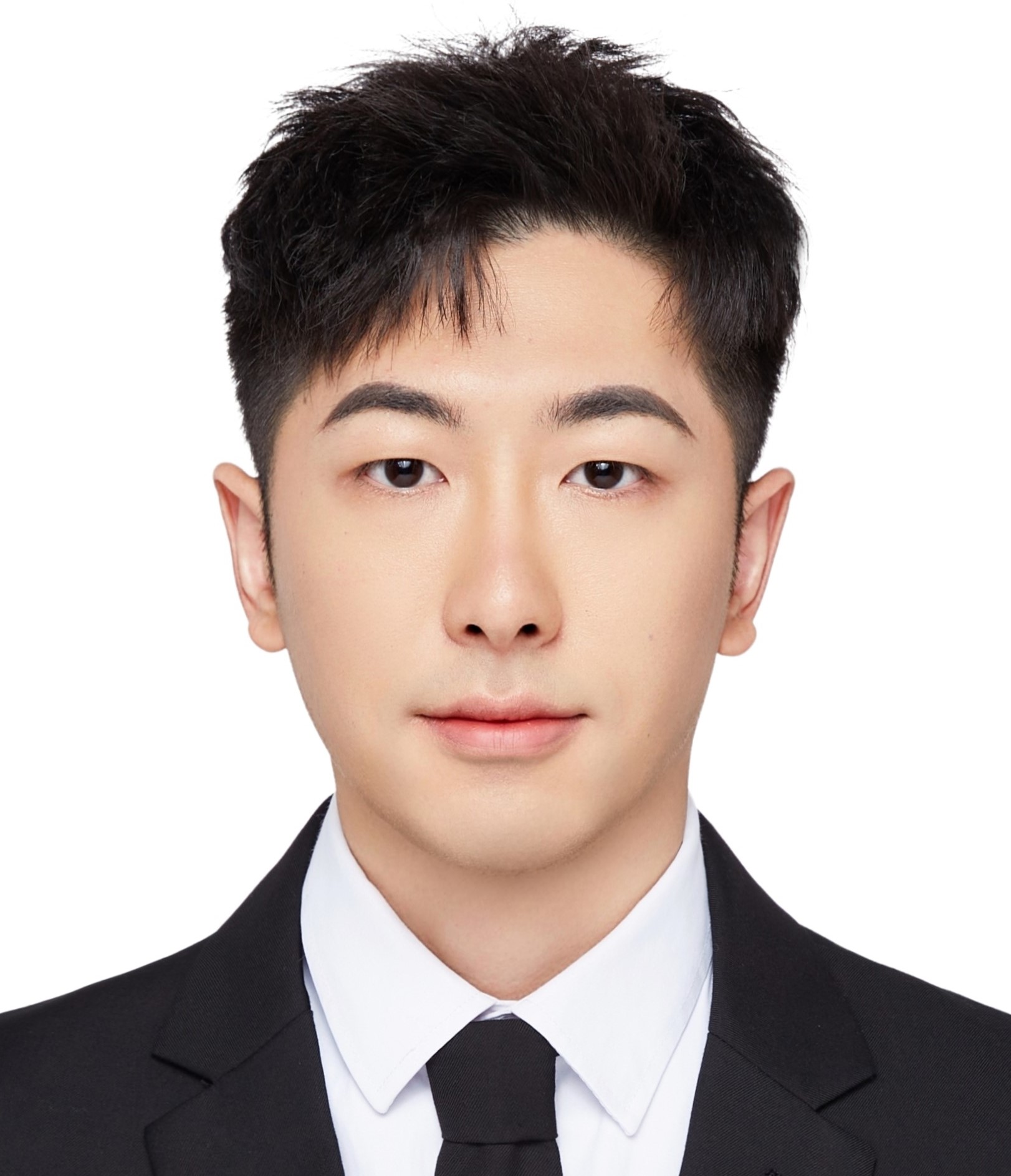}}]{Yao Wang} received the bachelor degree in aircraft manufacture engineering from the Changzhou Institute of Technology, Changzhou, China, in 2021. He is currently pursuing the master degree in mechanical engineering from the School of Mechanical and Electrical Engineering, Soochow University, Suzhou, China.
 
    His research field include mobile robot social navigation and motion planning.
\end{IEEEbiography}

\vspace{-0.4cm}

\begin{IEEEbiography}
	[{\includegraphics[width=1in,height=1.25in,clip,keepaspectratio]{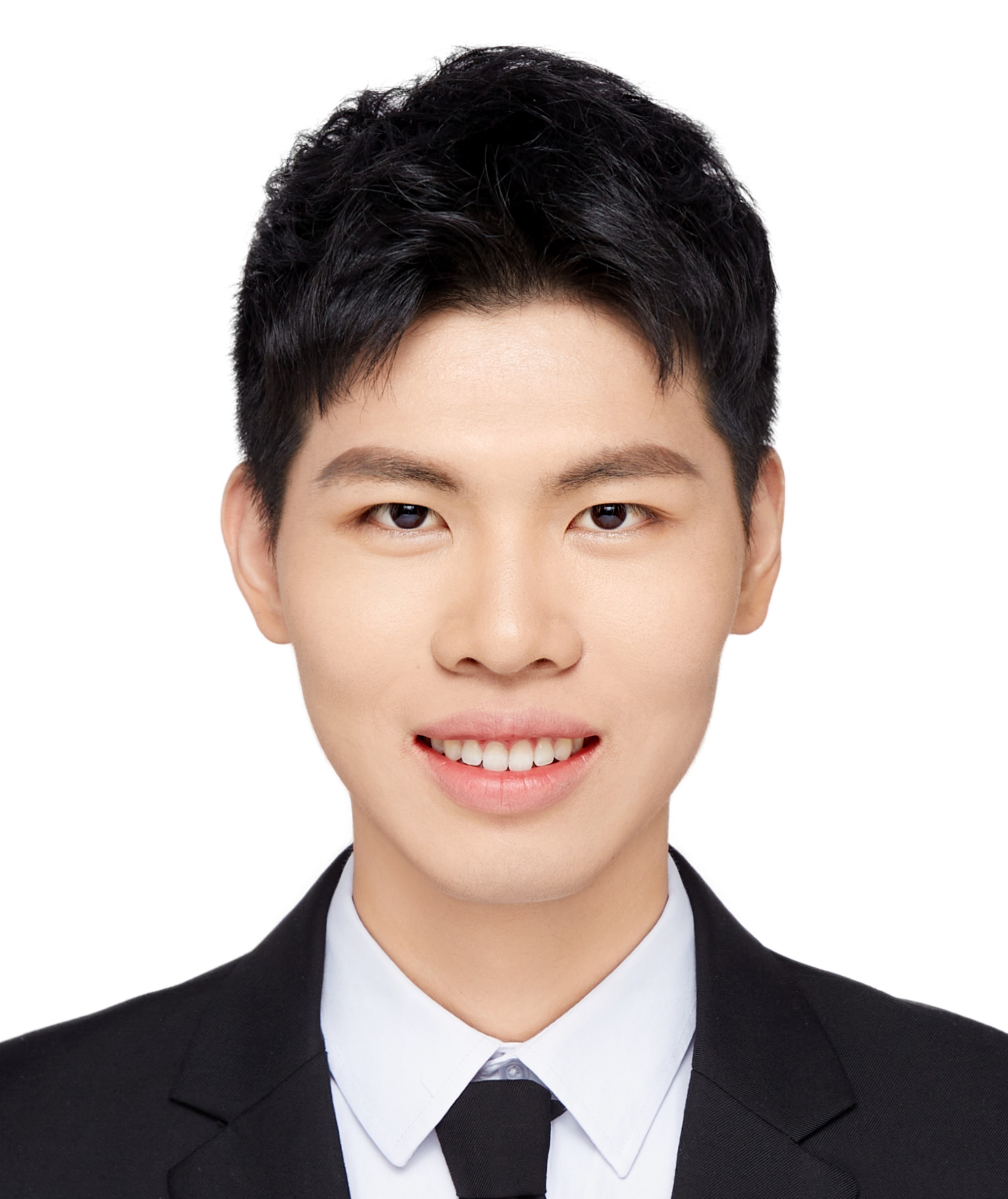}}]{Yuqi Kong} received the bachelor degree in mechanical and electronic engineering from the Changshu Institute of Technology, Suzhou, China, in 2021. He is currently pursuing the master degree in mechanical engineering from the School of Mechanical and Electrical Engineering, Soochow University, Suzhou, China.
 
    His research field include mobile robot social navigation and motion planning.
\end{IEEEbiography}

\vspace{-0.4cm}

\begin{IEEEbiography}
	[{\includegraphics[width=1in,height=1.25in,clip,keepaspectratio]{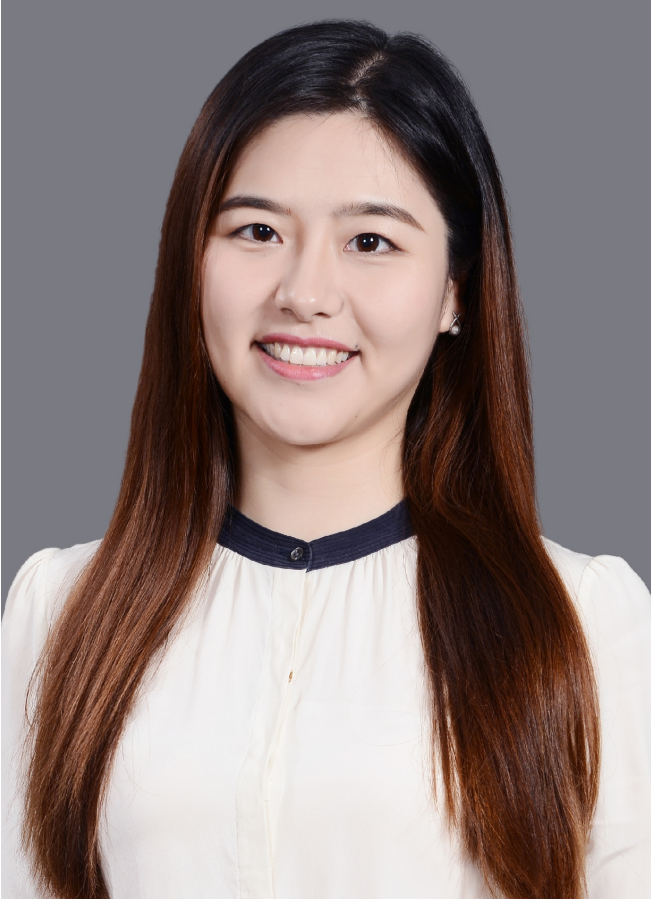}}]{Wenzheng Chi} received her B.E. degree in automation from Shandong University, Jinan, China, in 2013 and the Ph.D. degree in Biomedical Engineering from the Department of Electronic Engineering, The Chinese University of Hong Kong, Hong Kong, in 2017. 
 
    During which she spent six months at The University of Tokyo, Japan, as a Visiting Scholar. She was a Post-Doctoral Fellow with the Department of Electronic Engineering, The Chinese University of Hong Kong, Hong Kong, from 2017 to 2018. She is currently an Associate Professor with the Robotics and Microsystems Center, School of Mechanical and Electric Engineering, Soochow University, Suzhou, China. Her research interests include mobile robot path planning, intelligent perception, human-robot interaction, etc. 
\end{IEEEbiography}

\begin{IEEEbiography}
	[{\includegraphics[width=1in,height=1.25in,clip,keepaspectratio]{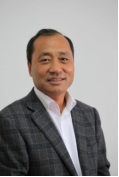}}]{Lining Sun} was born in Hegang, Heilongjiang Province, China, in January 1964. He received the Ph.D. degree in engineering from the Mechanical Engineering Department, Harbin Institute of Technology (HIT), Harbin, China, in 1993. 
 
    After graduation, he joined HIT. He formed an internationally competitive robotics team and made outstanding contributions to the creation and development of robot-related disciplines in China.
 
    Dr. Sun is a National Outstanding Youth Fund Winner and a Changjiang Scholar Distinguished Professor by the Ministry of Education.
\end{IEEEbiography}

\end{document}